\documentclass[runningheads]{llncs}

\usepackage[T1]{fontenc}
\def\doi#1{\href{https://doi.org/\detokenize{#1}}{\url{https://doi.org/\detokenize{#1}}}}
\usepackage{graphicx}
\usepackage{listings}
\usepackage{algorithm}
\usepackage{algorithmic}

\usepackage{amsmath,amssymb}
\usepackage{cleveref}
\usepackage{subcaption}
\usepackage{stmaryrd}
\usepackage{booktabs}
\usepackage{dirtytalk}
\usepackage{multirow}
\usepackage{adjustbox}

\usepackage[acronym]{glossaries}
\newacronym{es}{ES}{End-Systolic}
\newacronym{ed}{ED}{End-Diastolic}
\newacronym{ef}{LVEF}{Left Ventricular Ejection Fraction}
\newacronym{sv}{SV}{Systolic Volume}
\newacronym{us}{US}{Ultrasound}
\newacronym{vqvae}{VQ-VAE}{Vector Quantized-Variational AutoEncoder}
\newacronym{gan}{GAN}{Generative Adversarial Network}
\newacronym{dtgn}{D'ARTAGNAN}{Deep ARtificial Twin-Architecture GeNerAtive Networks}
\begin{document}

\title{D'ARTAGNAN: Counterfactual Video Generation}
\titlerunning{D'ARTAGNAN: Counterfactual Video Generation}

\author{
Hadrien Reynaud\inst{1,2} \and Athanasios~Vlontzos\inst{2} \and Mischa~Dombrowski\inst{3} \and Ciarán~Gilligan-Lee\inst{4} \and Arian~Beqiri\inst{5,6} \and Paul~Leeson\inst{5,7} \and Bernhard~Kainz\inst{2,3}
}

\institute{
UKRI CDT in AI for Healthcare, Imperial College London, London, UK \email{hadrien.reynaud19@imperial.ac.uk} \and
Department of Computing, Imperial College London, London, UK \and
Friedrich--Alexander University Erlangen--N\"urnberg, DE \and
Spotify \& 
University College London, London, UK \and
Ultromics Ltd, Oxford, UK \and 
King’s College London, School of BioEng \& Imaging Sciences, London, UK \and
John Radcliffe Hospital, Cardiovascular Clinical Research Facility, Oxford, UK
}

\authorrunning{H. Reynaud et al.}

\maketitle              
\begin{abstract}
Causally-enabled machine learning frameworks could help clinicians to identify the best course of treatments by answering counterfactual questions. We explore this path for the case of echocardiograms by looking into the variation of the Left Ventricle Ejection Fraction, the most essential clinical metric gained from these examinations. We combine deep neural networks, twin causal networks and generative adversarial methods for the first time to build D'ARTAGNAN (Deep ARtificial Twin-Architecture GeNerAtive Networks), a novel \emph{causal} generative model. We demonstrate the soundness of our approach on a synthetic dataset before applying it to cardiac ultrasound videos to answer the question: "What would this echocardiogram look like if the patient had a different ejection fraction?". To do so, we generate new ultrasound videos, retaining the video style and anatomy of the original patient, while modifying the Ejection Fraction conditioned on a given input. We achieve an SSIM score of 0.79 and an R2 score of 0.51 on the counterfactual videos. Code and models are available at: \url{https://github.com/HReynaud/dartagnan}.
\end{abstract}
\section{Introduction}
How would this patient's scans look if they had a different \gls{ef}? How would this \gls{us} view appear if I turned the probe by 5 degrees? These are important causality related questions that physicians and operators ask explicitly or implicitly during the course of an examination in order to reason about the possible pathologies of the patient. In the second case, the interventional query of turning the probe is easy to resolve, by performing the action. However, in the first case, we ask a counterfactual question which cannot be answered directly. Indeed, it falls under the third and highest rung of Pearl's~\cite{Pearl2009} hierarchy of causation. 

Counterfactual queries probe into alternative scenarios that might have occurred had our actions been different. For the first question of this paper, we ask ourselves how the patient's scans would look if they had a different \gls{ef}. Here, the treatment would be the different ejection fraction, and the outcome, the different set of scans. Note that this is a query that is \emph{counter-to} our observed knowledge that the patients scans exhibited a specific \gls{ef}. As such, standard Bayesian Inference, which conditions on the observed data without any further considerations, is not able to answer this type of question. 

\setcounter{footnote}{0} 

\noindent\textbf{Related works}: 
Generating synthetic \gls{us} images can be performed with physics-based simulators \cite{shams2008real,cong2013fast,Mattausch2014,gao2009fast,burger2012real} and other techniques, such as registration-based methods \cite{ledesma2005spatio}. However, these methods are usually very computationally expensive and do not generate fully realistic images. With the shift toward deep learning, \gls{gan}-based techniques have emerged. They can be based on simulated \gls{us} priors or other imaging modalities (MRI, CT)~\cite{CRONIN2020105583,9110573,Tiago2021,Amirrajab2020,Samaneh2020,Tomar2021} to condition the anatomy of the generated \gls{us} images. 
Recently, many works explore machine learning as a tool to estimate interventional conditional distributions~\cite{yoon2018ganite,kocaoglu2018causalgan,louizos2017causal,assaad2021counterfactual}. However, fewer works focus on the counterfactual query estimation. \cite{pawlowski2020deep,oberst2019counterfactual} explore the Abduction-Action-Prediction paradigm and use deep neural networks for the abduction step, which is computationally very expensive. \cite{cuellar2020non} derive a parametric mathematical model for the estimation of one of the probabilities of causation, while \cite{vlontzos2021estimating} use \cite{balke1994counterfactual} to develop deep twin networks. 
The computer vision field also has a lot of interest in conditional generation problems. \cite{wang2020imaginator,tulyakov2018mocogan} perform conditional video generation from a video and a discrete class. \cite{ding2020ccgan} uses an image and a continuous value as input to produce a new image. \cite{sauer2021counterfactual,kocaoglu2018causalgan} introduce causality in their generation process, to produce images from classes.

\noindent\textbf{Contributions}: In this paper
(1) We extend the causal inference methodology known as Deep Twin Networks~\cite{vlontzos2021estimating} into a novel generative modelling method (\gls{dtgn} \footnote{\footnotesize D'Artagnan is the  fourth Musketeer from the French tale ``The three Musketeers''.}) able to handle counterfactual queries. 
(2) We apply our framework on the synthetic MorphoMNIST \cite{castro2019morphomnist} and real-world EchoNet-Dynamic \cite{Ouyang2020} datasets, demonstrating that our method can perform well in both fully controlled environments and on real medical cases.
To the best of our knowledge, this is an entirely novel approach and task, and thus the first time such an approach is explored for medical image analysis and computer vision.
Our work differentiates itself from all other generative methods by combining video generation with continuous conditional input in a new causal framework. This setup supports counterfactual queries to produce counterfactual videos using a semi-supervised approach which allows most standard labelled datasets to be used.

\section{Preliminaries}

\textbf{Structural causal models} \label{Section: Structural causal models}
We work in the Structural Causal Models (SCM) framework. Chapter $7$ of \cite{Pearl2009} gives an in-depth discussion. For an up-to-date, review of counterfactual inference and Pearl's Causal Hierarchy, see \cite{bareinboim20201on}.

As a brief reminder, we define a structural causal model (SCM) as a set of latent noise variables $U=\{u_1,\dots,u_n\}$ distributed as $P(U)$, a set of observable variables $V=\{v_1,\dots, v_m\}$ that is the superset of treatment variables $X$ and confounders $Z$, \emph{i.e.} $X,Z \subseteq V $. Moreover, we require  a directed acyclic graph (DAG), called the \emph{causal structure} of the model, whose nodes are the variables $U\cup V$, and its edges represent a collection of functions $F=\{f_1,\dots, f_n\}$, such that $v_i = f_i(PA_i, u_i), \text{ for } i=1,\dots, n,$ where $PA$ denotes the parent observed nodes of an observed variable. These are used to induce a distribution over the observable variables and assign uncertainty over them.

Finally, the $do$-operator forces variables to take certain values, regardless of the original causal mechanism. Graphically, $do(X=x)$ means deleting edges incoming to $X$ and setting $X=x$. Probabilities involving $do(x)$ are normal probabilities in submodel $M_x$: $P(Y=y \mid \text{do}(X=x)) = P_{M_x} (y)$.

\noindent\textbf{Counterfactual inference} \label{Section: two approaches to counterfactual inference}
The latent distribution $P(U)$ allows us to define probabilities of counterfactual queries, $P(Y_{y}=y) = \sum_{u \mid Y_{x}(u)=y} P(u)$. Moreover, one can define a counterfactual distribution given seemingly contradictory evidence and thereby state the counterfactual sentence ``$Y$ would be $y$ (in situation $U=u$), had $X$ been $x$''. Despite the fact that this query may involve interventions that contradict the evidence, it is well-defined, as the intervention specifies a new submodel. Indeed, $ P(Y_{x}={y}' \mid E = {e})$ is given by ~\cite{Pearl2009} as 
$ \sum_{{u}}
P(Y_{{x}}({u})={y}')P({u}|{e})\,.$

\noindent\textbf{Twin Networks}
As opposed to the standard \textit{Abduction - Action - Prediction} paradigm, we will be operating under the Twin Model methodology. Originally proposed by Balke and Pearl in \cite{balke1994counterfactual}, this method allows efficient counterfactual inference to be performed as a feed forward Bayesian process. It has also been shown empirically to offer computational savings relative to abduction-action-prediction~\cite{grahamcopy}. A twin network consists of two interlinked networks, one representing the real world and the other the counterfactual world being queried. 

Given a structural causal model, a twin network can be constructed and used to compute a counterfactual query through the following steps: First, duplicate the given causal model, denoting nodes in the duplicated model via superscript $^*$. Let $V $ be observable nodes that include the treatment variables $X$ and the confounders $Z$, and let $X^*,Z^* \subseteq V^{*}$ be the duplication of these. Also let $U$ be the unobserved latent noise.  Then, for every node $v_i^{*}$ in the duplicated, or ``counterfactual'' model, its latent parent $u_i^{*}$ is replaced with the original latent parent $u_i$ in the original, or ``factual'', model, such that the original latent variables are now a parent of two nodes, $v_i$ and $v_i^{*}$. The two graphs are linked only by common latent parents, but share the same node structure and generating mechanisms. To compute a general counterfactual query $P(Y=y \mid E=e, \text{do}(X=x))$, modify the structure of the counterfactual network by dropping arrows from parents of $X^*$ and setting them to value $X^*=x$. Then, in the twin network with this modified structure, compute the probability $P(Y^*=y \mid E=e, X^*=x)$ via standard Bayesian inference techniques, where $E$ are factual nodes.

\section{Method} \label{sec:dtgn}

\begin{figure}[t]
    \centering
   {
        \includegraphics[width=0.95\linewidth]{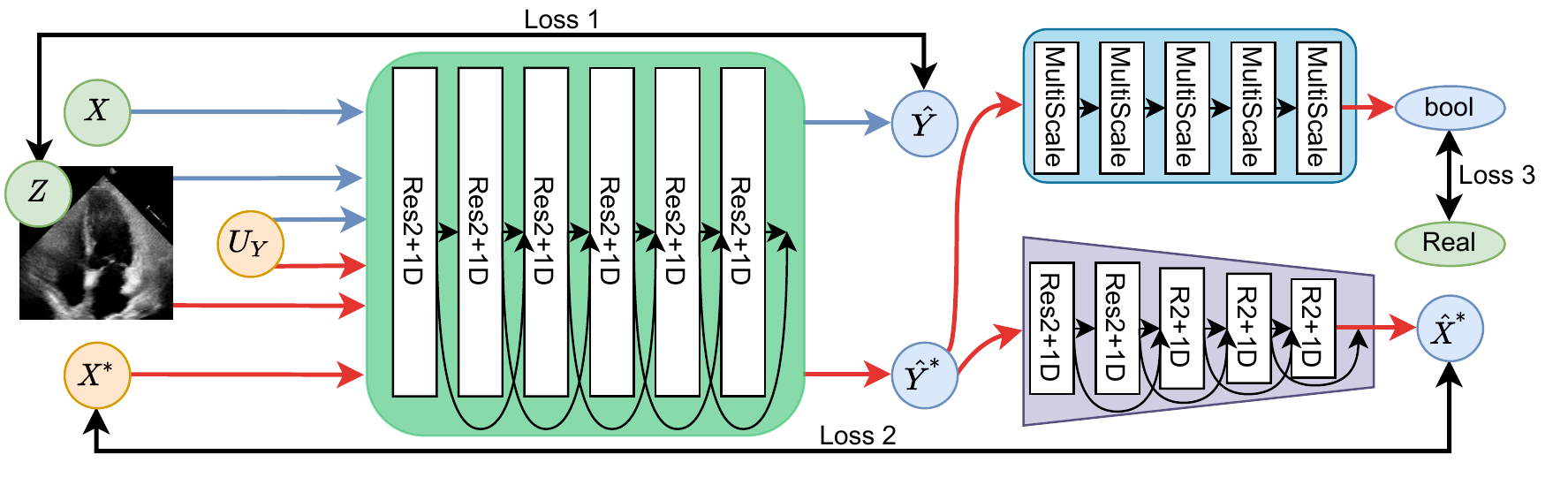}
        }
    \caption{
    The \gls{dtgn} framework. The green variables are known, the orange are sampled from distributions and the blue are generated by deep neural networks. Factual path is in blue, counterfactual in red. \label{figure: deep twin netork architectures_TN}}
\end{figure}

\noindent\textbf{Deep Twin Networks}
The methodology we propose is based on Deep Twin Networks. The training procedure and parametrization are borrowed from \cite{vlontzos2021estimating}, which sets the foundation for our causal framework. Deep Twin Networks use two branches, the factual and the counterfactual branch. We denote our factual and counterfactual treatments (inputs unique to each branch) as $X$ and $X^*$, while the confounder (input shared between both branches) is denoted by $Z$. We denote the factual and counterfactual outcomes as $\hat{Y}, \hat{Y}^*$ while the noise, injected midway through the network, and shared by both branches, is denoted by $U_Y$. This information flow sets $Z$ as the data we want to query with $X$ and $X^*$, to produce the outcomes $\hat{Y}$ and $\hat{Y}^*$. These variables will be detailed on a case-specific basis in \Cref{sec:experiments}. See Fig.\ref{figure: deep twin netork architectures_TN} for a visual representation of the information flow.

\noindent\textbf{Synthetic data}
Synthetic data allows for full control over the generation of both ground truth outcomes $Y$ and $Y^*$ along with their corresponding inputs $X$, $X^*$ and $Z$. This makes training the Deep Twin Network trivial in a fully supervised fashion, as demonstrated in our first experiment.

\noindent\textbf{Real-world data}
Theoretically, our approach requires labelled data pairs, but very few datasets are arranged as such. To overcome this limitation and support most standard labeled imaging datasets, we establish a list of features that our model must possess to generate counterfactual videos for the medical domain: (1) produce a factual and counterfactual output that share general visual features, such as style and anatomy, (2) produce accurate factual and counterfactual videos with respect to the treatment variable, and (3) the produced videos must be visually indistinguishable from real ones. In the following, we use the Echonet-Dynamic~\cite{Ouyang2020} dataset to illustrate the method, see \Cref{sec:experiments} for details. 
We solve feature (1) by sharing the weights of the branches in the network, such that we virtually train a single branch on two tasks in parallel. To do so, we set the confounder $Z$ as the input video and the treatment $X$ as the labelled \gls{ef}. We train the network to match the factual outcome $\hat{Y}$ with the input video. By doing so, the model learns to retain the style and anatomical structure of the echocardiogram from the confounder. This is presented as \emph{Loss 1} in \Cref{figure: deep twin netork architectures_TN}.
For feature (2) we pre-train an expert network to regress the treatment values from the videos produces by the counterfactual branch, and compare them to the counterfactual treatment. The expert network's weights are frozen when training the Twin model, and alleviates the need for labels to train the counterfactual branch. This loss is denoted as \emph{Loss 2} in \Cref{figure: deep twin netork architectures_TN}.
Finally, feature (3) calls for the well-known \gls{gan} framework, where we train a neural network to discriminate between real and fake images or videos, while training the Twin Network. This constitutes the adversarial \emph{Loss 3} in \Cref{figure: deep twin netork architectures_TN} and ensures that the counterfactual branch produces realistic-looking videos.

With those 3 losses, we can train the generator (i.e. factual and counterfactual branches) to produce pairs of visually accurate and anatomically matching videos that respect their factual and counterfactual \gls{ef}s treatment.
To learn the \textbf{noise distribution} $U_Y$, we follow~\cite{goudet2017learning,vlontzos2021estimating} and without loss of generality we can write $Y=f(X,Z,g(U_Y'))$ with $U_Y'\sim \mathcal{E}$ and $U_Y = g(U_Y')$, where $\mathcal{E}$ is some easy-to-sample-from distribution, such as a Gaussian or Uniform. Effectively, we cast the problem of determining $U_Y$ to learning the appropriate transformation from $\mathcal{E}$ to $U_Y$. For ease of understanding, we will be using $U_Y$ henceforth to signify our approximation $g(U_Y')$ of the true unobserved parameter $U_Y$.
In addition to specifying the causal structure, the following standard assumptions are needed to correctly estimate $\mathbb{E}(Y | do(X), Z)$ \cite{schwab2018perfect}: (1) \emph{Ignorability:} there are no unmeasured confounders; (2) \emph{Overlap:} every unit has non-zero probability of receiving all treatments given their observed covariates.

\section{Experimentation}\label{sec:experiments}
\noindent\textbf{Datasets} 
To evaluate \gls{dtgn}, we use two publicly available datasets, the synthetic MorphoMNIST~\cite{castro2019morphomnist} and the clinical Echonet-Dynamic~\cite{Ouyang2020} dataset.

MorphoMNIST is a set of tools that enable fine-grained perturbations of the MNIST digits through four morphological functions, as well as five measurements of the digits. To train our model, we need five elements: an original image, a (counter-)factual treatment $X$($X^*$) and a corresponding (counter-)factual label $Y$($Y^*$). To generate this data, we take 60,000 MNIST images $I_i$ and sample 40 perturbation vectors $p_{i,j}$ for the five possible perturbations, including identity, thus generating 2.4 million images $I_{p_{i,j}}$. The perturbation vectors also encode the relative positions of the perturbations, when applicable. We \textit{measure} the original images to produce vectors $m_i$ and one-hot encode the labels into vectors $l_i$. We decided to perform the causal reasoning over a latent space, rather than image space. To do so, we train a \gls{vqvae} \cite{oord2017neural} to project the MorphoMNIST images to a latent space $\mathcal{H} \in \mathbb{R}^{(q \times h \times w)} $ and reconstruct them. Once trained, the \gls{vqvae} weights are frozen, and we encode all the ground-truth perturbed images $I_{p_{i,j}}$ into a latent embedding $\mathcal{H}_{i,j}$. Afterwards, the \gls{vqvae} is used to reconstruct the generated latent embeddings $\hat{Y}, \hat{Y}^*$ for qualitative evaluation purposes.

The clinical dataset Echonet-Dynamic~\cite{Ouyang2020} consists of 10,030 4-chamber echo\-cardiography videos with $112 \times 112$ pixels resolution and various length, frame rates, image quality and cardiac conditions. Each video contains a single pair of consecutively labelled \gls{es} and \gls{ed} frames. Each video also comes with a manually measured \gls{ef}. For our use case, all videos are greyscaled and resampled to 64 frames, with a frame rate of 32 images per second. All videos shorter than two seconds are discarded, and we make sure to keep the labelled frames. For the resulting 9724 videos dataset, the original split is kept, with 7227 training, 1264 validation and 1233 testing videos.

\begin{figure}[!t]
\centering

  \centering
  \includegraphics[width=0.12\linewidth]{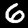}~~~~
  \includegraphics[width=0.12\linewidth]{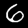}
  \includegraphics[width=0.12\linewidth]{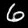}~~~~
  \includegraphics[width=0.12\linewidth]{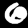}
  \includegraphics[width=0.12\linewidth]{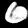}

\caption{Left to right: original image, GT factual image, predicted factual image, GT counterfactual image, predicted counterfactual image. Factual perturbation is Thinning, Counterfactual perturbation are Thickening and Swelling.}
\label{fig:mnist_images}

\end{figure}

\noindent\textbf{MorphoMNIST} For our synthetic experiment, we define a deep twin network as in~\cite{vlontzos2021estimating} and follow the process we described in the Methods (\Cref{sec:dtgn}). Regarding data organization, we use the elements defined in the section above. We set our confounder $Z_i = [l_i, m_i]$ to contain the one-hot encoded labels as well as the measurement of the original image. We sample two perturbation vectors $p_{i,m}$ and $p_{i,n}$ and their corresponding latent embeddings $\mathcal{H}_{i,m}$ and $\mathcal{H}_{i,n}$, where $n, m \in \llbracket 0, 40\llbracket$, $n \neq m$. We set our input treatments as the perturbations vectors ($X = p_{i,m}$, $X^* = p_{i,n}$) and our ground-truth outcomes as the corresponding latent embeddings of the perturbed images ($Y=\mathcal{H}_{i,m}$, $Y^*=\mathcal{H}_{i,n}$). We sample $U_Y \sim [ \mathcal{N}(0 ,0.25)\bmod{1} + 1 ]$ and disturb the output of the branches of the neural networks that combine $X$ and $X^*$ with $Z$ by multiplying the outputs of both with the same $U_Y$. With this setup, we generate factual and counterfactual perturbed MNIST embeddings from a latent description. 

We assess the quality of the results by three means:
(1) \emph{Embeddings' MSE:} We sample a quintuplet ($Z$, $X$, $X^*$, $Y$, $Y^*$) and 1000 $U_Y$. The MSE between all $\hat{Y}_{i}$ and $Y$ are computed and used to order the pairs ($\hat{Y}_i$, $\hat{Y}^*_i$) in ascending order. We keep the sample with the lowest MSE as our factual estimate and compute the MSE between $\hat{Y}^*_0$ and $Y^*$ to get our counterfactual MSE score.
(2) \emph{SSIM:} We use the Structural SIMilarity metric between the perturbed images $I_{gt} = I_{p_{i,j}}$, the images reconstructed by the VQVAE $I_{rec}$ and the images reconstructed from the latent embedding produced by the twin network $I_{pred}$ to get a quantitative score over the images. 
(3) \emph{Images:} We sample some images to qualitatively assess best and worst cases scenarios for this framework.
We show the quantitative results in \Cref{tab:metrics}(a), and qualitative results in~\Cref{fig:mnist_images} and in the appendix.

\begin{table}[t]
\centering
\caption{Metrics for MorphoMNIST (a) and EchoNet-Dynamic (b) experiments.}
\begin{subtable}{.6\textwidth}
      \centering
    \begin{tabular}{|l|cc|}
        \hline
        Metric & Factual & Counterfactual\\
        \hline
        MSE($Y$, $\hat{Y}$)                 & 2.3030 & 2.4232 \\
        SSIM($I_{gt}$, $I_{rec}$)$^\dagger$ & 0.9308 & 0.9308 \\
        SSIM($I_{rec}$, $I_{pred}$)         & 0.6759 & 0.6759 \\
        SSIM($I_{gt}$, $I_{pred}$)          & 0.6707 & 0.6705 \\
        \hline
    \end{tabular}
    \caption{MSE and SSIM scores. $^\dagger$No ordering is performed as there is no noise involved. \label{tab:mnist_metrics}}
\end{subtable}
\begin{subtable}{.39\textwidth}
      \centering
    \begin{tabular}{|l|cc|}
        \hline
        Metric  & Factual   & Counterf. \\
        \hline
        R2      & 0.87      & 0.51 \\
        MAE     & 2.79      & 15.7 \\
        RMSE    & 4.45      & 18.4 \\
        SSIM    & 0.82      & 0.79 \\
        \hline
    \end{tabular}
    \caption{\gls{dtgn} LVEF and reconstruction metrics.\label{tab:gan_metrics}}
\end{subtable}
\label{tab:metrics}

\end{table}

\noindent\textbf{Echonet Dynamic}
As stated in~\Cref{sec:dtgn}, our methodology requires an \textbf{Expert Model} to predict the \gls{ef} of any \gls{us} video. To do so, we re-implement the ResNet 2+1D network~\cite{tran2018closer} as it was shown to be the best option for \gls{ef} regression in \cite{Ouyang2020}. We opt not to use transformers as they do not supersede convolutions for managing the temporal dimension, as shown in \cite{reynaud2021ultrasound}. 
We purposefully keep this model as small as possible in order to minimize its memory footprint, as it will be operating together with the generator and the frame discriminator. The expert network is trained first, and frozen while we train the rest of \gls{dtgn}. Metrics for this model are presented in the Appendix. 

\noindent\textbf{\gls{dtgn}}
We implement the generator as described in \Cref{sec:dtgn}. We define a single deep network to represent both the factual and counterfactual paths. By doing so, we can meet the objectives listed in \Cref{sec:dtgn}. The branch is implemented as a modified ResNet 2+1D \cite{tran2018closer} to generate videos. It takes two inputs: a continuous value and a video, where the video determines the size of the output. For additional details, please refer to \Cref{figure: deep twin netork architectures_TN} and the code.

\noindent\textbf{Discriminator}
We build a custom discriminator architecture using five ``residual multiscale convolutional blocks", with kernel sizes 3, 5, 7 and appropriate padding at each step, followed by a max-pooling layer. Using multiscale blocks enables the discriminator to look at both local and global features. This is extremely important in \gls{us} images because of the noise in the data, that needs to be both ignored, for anatomical identification, and accounted for to ensure that counterfactual \gls{us} images look real. We test this discriminator both as a frame-based discriminator and a video-based discriminator, by changing the 2D layers to 3D layers where appropriate. We note that, given our architecture, the 3D version of the model requires slightly less memory but doubles the processing power compared to the 2D model. 

\noindent\textbf{Training the framework}
At each training step, we sample an \gls{us} video ($V$) and its \gls{ef} ($\psi$). We set our factual treatment, $X = \psi$ and our counterfactual treatment $X^* \sim \mathcal{U}(0,\psi-0.1) \cup \mathcal{U}(\psi+0.1,1)$. The confounder $Z$ and the factual ground truth $Y$ are set to the sampled video such that $Z = Y = V$. We compute an L1 reconstruction loss (loss 1) between $\hat{Y}$ and $Y = V$. As we do not have ground truth for the counterfactual branch, we use the frame discriminator and the expert model to train it. Both models take as input the counterfactual prediction $\hat{Y}^*$. The expert model predicts an \gls{ef} $\hat{\psi}$ that is trained to match the counterfactual input $X^*$ with an L1 loss (loss 2). The discriminator is trained as a \gls{gan} discriminator, with $\hat{Y}^*$ as fake samples and $V$ as real samples. It trains the generator with L1 loss (loss 3).
The discriminator and expert model losses are offset respectively by three and five epochs, leaving time for the generator to learn to reconstruct $V$, thus maintaining the anatomy and style of the confounder. Our experiments show that doing so increases the speed at which the network is capable of outputting realistic-looking videos, thus speeding up the training of the discriminator that sees accurate fake videos sooner. Once the generator and discriminator are capable of generating and discriminating realistic-looking videos, the expert network loss is activated and forces the generator to take into account the counterfactual treatment, while the factual treatment is enforced by the reconstruction loss. The losses are also scaled, such that the discriminator loss has a relative weight of 3 compared to the reconstruction and expert loss.

\begin{figure}[!t]
    \centering
    \begin{minipage}{.35\textwidth}
    \centering
      \begin{tabular}{cc}
        \includegraphics[trim={0 1cm 0 1cm},clip,width=0.5\linewidth]{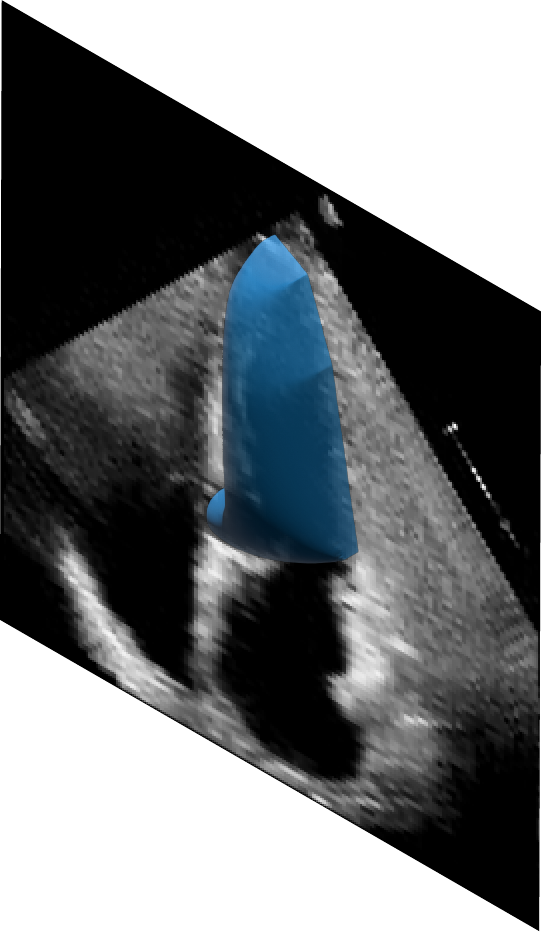} &
        \includegraphics[trim={0 1cm 0 1cm},clip,width=0.5\linewidth]{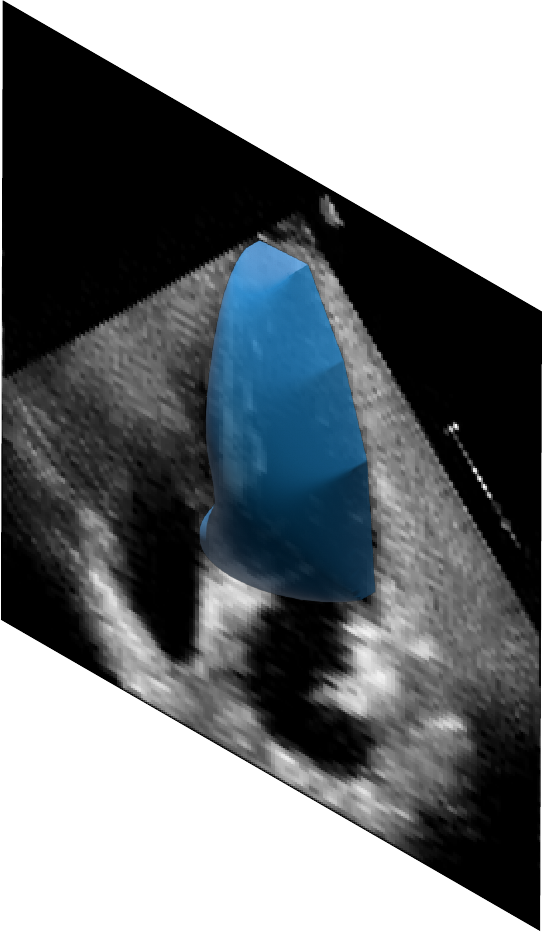} \\
        Factual \gls{es} & Factual \gls{ed} \\
      \end{tabular}
    \end{minipage}
    \begin{minipage}{.29\textwidth}
    \centering
      \hfill\includegraphics[width=0.95\linewidth]{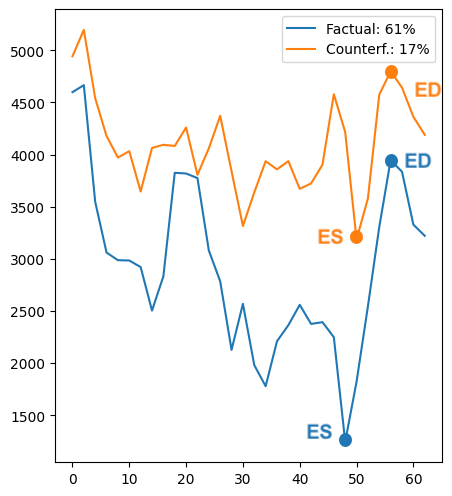}
    \end{minipage}
    \begin{minipage}{.35\textwidth}
    \centering
      \begin{tabular}{cc}
        \includegraphics[trim={0 1cm 0 1cm},clip,width=0.5\linewidth]{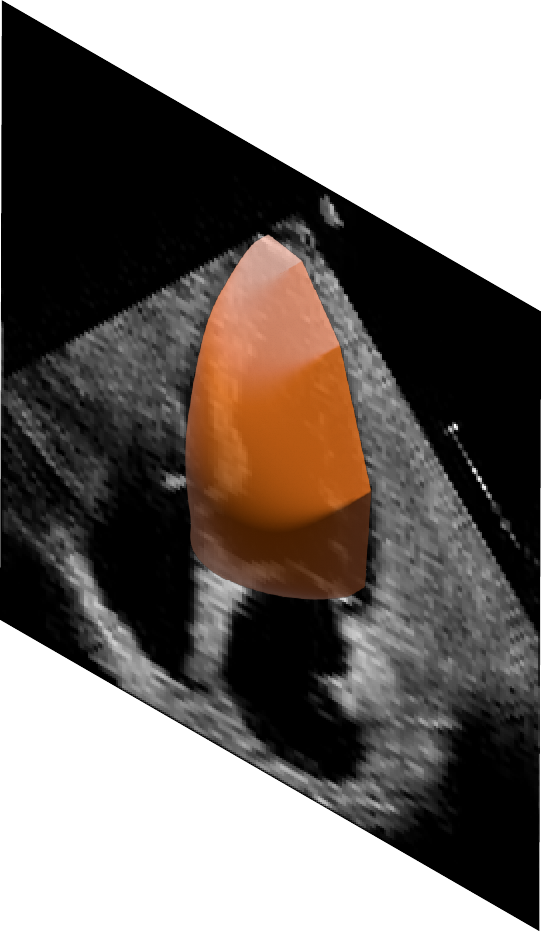} &
        \includegraphics[trim={0 1cm 0 1cm},clip,width=0.5\linewidth]{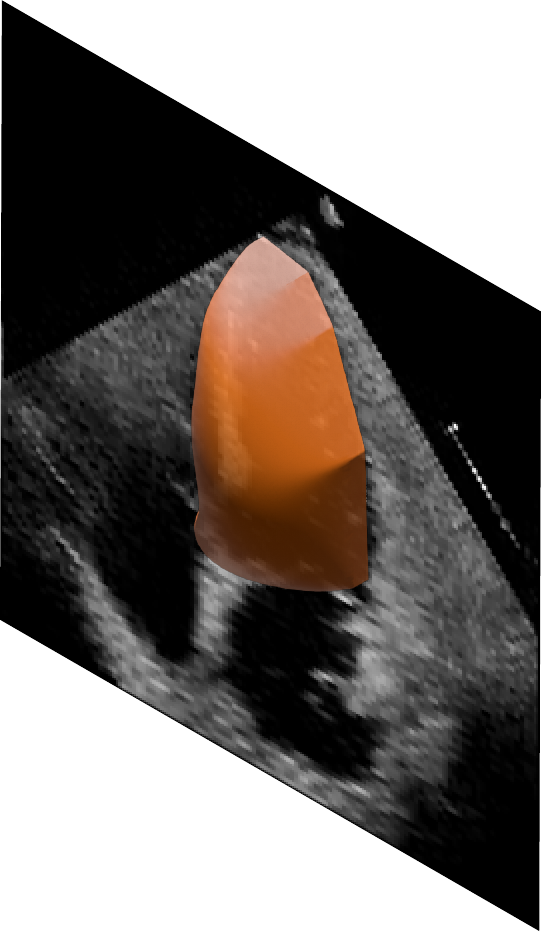} \\
        Counterf \gls{es} & Counterf. \gls{ed} \\
      \end{tabular}
    \end{minipage}
    \caption{Qualitative results for \gls{dtgn} over the same confounder and noise. Left: factual \gls{es} and \gls{ed} frames. Middle: left ventricle area over time, obtained with a segmentation network as in \cite{Ouyang2020}. Dots represent where the corresponding frames were sampled. Right: counterfactual \gls{es} and \gls{ed} frames.
    Anatomy is preserved across videos, while the \gls{ef} fraction is different. 
    }
    \label{fig:quali_echo}
\end{figure}

\noindent\textbf{Metrics}
To match our three objectives, we evaluate this new task on 1) the accuracy of the anatomy and 2) the accuracy of the regressed \gls{ef}. 
We obtain the best possible video by sampling 100 $U_Y$ and keeping the $\hat{Y}^*$ with $\hat{\phi}^*$ closest to $X^*$.
We then evaluate our metrics over those ``best'' videos. The anatomical accuracy is measured with SSIM and the \gls{ef} precision is measured using R2, MAE and RMSE scores. Results are shown in \Cref{tab:gan_metrics}.

\noindent\textbf{Qualitative} In \Cref{fig:quali_echo} we show video frame examples to showcase the quality of the reconstructed frames, as well as how the anatomy and style are maintained.

\noindent\textbf{Discussion} 
The predicted \gls{ef} has an MAE of 15.7\% which is not optimal. This can come from many factors such as the use of the same dataset to train the Expert model and \gls{dtgn}, the limited number of real videos, or the limited size of the networks due to the necessity of working with three models at once. Those problems could be addressed with hyperparameter search, larger models, as well as additional medical data. For completeness, we compare the performance of \gls{dtgn} with the literature~\cite{wang2020imaginator,tulyakov2018mocogan}, and run additional experiments (see Appendix) with an ablated version of our model for conditional video generation, where it achieves the best SSIM score of 0.72.

\section{Conclusion}
In this paper we introduce \gls{dtgn}, a Deep Twin Generative Network able to produce counterfactual images and videos. We showcase its performance in both a synthetic and a real world medical dataset and achieve visually accurate results and high quantitative scores. In future work, we will explore other treatments, such as changing the heartbeat, as well as less constrained confounders, e.g. image segmentations.\\

\noindent\textbf{Acknowledgements:} This work was supported by Ultromics Ltd., the UKRI Centre for Doctoral Training in Artificial Intelligence for Healthcare (EP/S023283/1), and the UK Research and Innovation London Medical Imaging and Artificial Intelligence Centre for Value Based Healthcare. We thank the NVIDIA corporation for their GPU donations used in this work.

\bibliographystyle{splncs04}
\bibliography{bibliography.bib}

\begin{thebibliography}{10}
\providecommand{\url}[1]{\texttt{#1}}
\providecommand{\urlprefix}{URL }
\providecommand{\doi}[1]{https://doi.org/#1}

\bibitem{Samaneh2020}
Abbasi-Sureshjani, S., Amirrajab, S., Lorenz, C., Weese, J., Pluim, J.,
  Breeuwer, M.: 4d semantic cardiac magnetic resonance image synthesis on xcat
  anatomical model. arXiv:2002.07089  (2 2020)

\bibitem{Amirrajab2020}
Amirrajab, S., Abbasi-Sureshjani, S., Khalil, Y.A., Lorenz, C., Weese, J.,
  Pluim, J., Breeuwer, M.: Xcat-gan for synthesizing 3d consistent labeled
  cardiac mr images on anatomically variable xcat phantoms. arXiv:2007.13408
  (7 2020)

\bibitem{assaad2021counterfactual}
Assaad, S., Zeng, S., Tao, C., Datta, S., Mehta, N., Henao, R., Li, F., Duke,
  L.C.: Counterfactual representation learning with balancing weights. In:
  International Conference on Artificial Intelligence and Statistics. pp.
  1972--1980. PMLR (2021)

\bibitem{balke1994counterfactual}
Balke, A., Pearl, J.: Probabilistic evaluation of counterfactual queries. In:
  AAAI (1994)

\bibitem{bareinboim20201on}
Bareinboim, E., Correa, J.D., Ibeling, D., Icard, T.: On pearl’s hierarchy
  and the foundations of causal inference. Tech. rep., Columbia University,
  Stanford University (2020)

\bibitem{burger2012real}
Burger, B., Bettinghausen, S., Radle, M., Hesser, J.: Real-time gpu-based
  ultrasound simulation using deformable mesh models. IEEE transactions on
  medical imaging  \textbf{32}(3),  609--618 (2012)

\bibitem{castro2019morphomnist}
Castro, D.C., Tan, J., Kainz, B., Konukoglu, E., Glocker, B.: {Morpho-MNIST}:
  Quantitative assessment and diagnostics for representation learning. Journal
  of Machine Learning Research  \textbf{20}(178) (2019)

\bibitem{cong2013fast}
Cong, W., Yang, J., Liu, Y., Wang, Y.: Fast and automatic ultrasound simulation
  from ct images. Computational and mathematical methods in medicine  (2013)

\bibitem{CRONIN2020105583}
Cronin, N.J., Finni, T., Seynnes, O.: Using deep learning to generate synthetic
  b-mode musculoskeletal ultrasound images. Computer Methods and Programs in
  Biomedicine  \textbf{196},  105583 (2020)

\bibitem{cuellar2020non}
Cuellar, M., Kennedy, E.H.: A non-parametric projection-based estimator for the
  probability of causation, with application to water sanitation in kenya.
  Journal of the Royal Statistical Society: Series A (Statistics in Society)
  \textbf{183}(4),  1793--1818 (2020)

\bibitem{ding2020ccgan}
Ding, X., Wang, Y., Xu, Z., Welch, W.J., Wang, Z.J.: Ccgan: Continuous
  conditional generative adversarial networks for image generation. In:
  International Conference on Learning Representations (2020)

\bibitem{gao2009fast}
Gao, H., Choi, H.F., Claus, P., Boonen, S., Jaecques, S., Van~Lenthe, G.H.,
  Van~der Perre, G., Lauriks, W., D'hooge, J.: A fast convolution-based
  methodology to simulate 2-dd/3-d cardiac ultrasound images. IEEE transactions
  on ultrasonics, ferroelectrics, and frequency control  \textbf{56}(2),
  404--409 (2009)

\bibitem{goudet2017learning}
Goudet, O., Kalainathan, D., Caillou, P., Lopez-Paz, D., Guyon, I., Sebag, M.,
  Tritas, A., Tubaro, P.: Learning functional causal models with generative
  neural networks. arXiv:1709.05321  (2017)

\bibitem{grahamcopy}
Graham, L., Lee, C.M., Perov, Y.: Copy, paste, infer: A robust analysis of twin
  networks for counterfactual inference. NeurIPS Causal ML workshop 2019,
  (2019)

\bibitem{kocaoglu2018causalgan}
Kocaoglu, M., Snyder, C., Dimakis, A.G., Vishwanath, S.: Causalgan: Learning
  causal implicit generative models with adversarial training. In:
  International Conference on Learning Representations (2018)

\bibitem{ledesma2005spatio}
Ledesma-Carbayo, M.J., Kybic, J., Desco, M., Santos, A., Suhling, M., Hunziker,
  P., Unser, M.: Spatio-temporal nonrigid registration for ultrasound cardiac
  motion estimation. IEEE transactions on medical imaging  \textbf{24}(9),
  1113--1126 (2005)

\bibitem{louizos2017causal}
Louizos, C., Shalit, U., Mooij, J., Sontag, D., Zemel, R., Welling, M.: Causal
  effect inference with deep latent-variable models. In: Proceedings of the
  31st International Conference on Neural Information Processing Systems. pp.
  6449--6459 (2017)

\bibitem{Mattausch2014}
Mattausch, O., Makhinya, M., Goksel, O.: Realistic ultrasound simulation of
  complex surface models using interactive monte-carlo path tracing (2014)

\bibitem{oberst2019counterfactual}
Oberst, M., Sontag, D.: Counterfactual off-policy evaluation with gumbel-max
  structural causal models. In: International Conference on Machine Learning.
  pp. 4881--4890. PMLR (2019)

\bibitem{oord2017neural}
Oord, A.v.d., Vinyals, O., Kavukcuoglu, K.: Neural discrete representation
  learning. arXiv:1711.00937  (2017)

\bibitem{Ouyang2020}
Ouyang, D., He, B., Ghorbani, A., Yuan, N., Ebinger, J., Langlotz, C.P.,
  Heidenreich, P.A., Harrington, R.A., Liang, D.H., Ashley, E.A., Zou, J.Y.:
  Video-based ai for beat-to-beat assessment of cardiac function. Nature
  \textbf{580},  252--256 (4 2020)

\bibitem{pawlowski2020deep}
Pawlowski, N., Castro, D.C., Glocker, B.: Deep structural causal models for
  tractable counterfactual inference. arXiv:2006.06485  (2020)

\bibitem{Pearl2009}
Pearl, J.: Causality (2nd edition). Cambridge University Press (2009)

\bibitem{reynaud2021ultrasound}
Reynaud, H., Vlontzos, A., Hou, B., Beqiri, A., Leeson, P., Kainz, B.:
  Ultrasound video transformers for cardiac ejection fraction estimation. In:
  MICCAI. pp. 495--505. Springer (2021)

\bibitem{sauer2021counterfactual}
Sauer, A., Geiger, A.: Counterfactual generative networks. arXiv:2101.06046
  (2021)

\bibitem{schwab2018perfect}
Schwab, P., Linhardt, L., Karlen, W.: Perfect match: A simple method for
  learning representations for counterfactual inference with neural networks.
  arXiv:1810.00656  (2018)

\bibitem{shams2008real}
Shams, R., Hartley, R., Navab, N.: Real-time simulation of medical ultrasound
  from ct images. In: International Conference on Medical Image Computing and
  Computer-Assisted Intervention. pp. 734--741. Springer (2008)

\bibitem{9110573}
Teng, L., Fu, Z., Yao, Y.: Interactive translation in echocardiography training
  system with enhanced cycle-gan. IEEE Access  \textbf{8},  106147--106156
  (2020)

\bibitem{Tiago2021}
Tiago, C., Gilbert, A., Snare, S.R., Sprem, J., McLeod, K.: Generation of 3d
  cardiovascular ultrasound labeled data via deep learning (2021)

\bibitem{Tomar2021}
Tomar, D., Zhang, L., Portenier, T., Goksel, O.: Content-preserving unpaired
  translation from simulated to realistic ultrasound images. arXiv:2103.05745
  (3 2021)

\bibitem{tran2018closer}
Tran, D., Wang, H., Torresani, L., Ray, J., LeCun, Y., Paluri, M.: A closer
  look at spatiotemporal convolutions for action recognition. In: Proceedings
  of the IEEE conference on Computer Vision and Pattern Recognition. pp.
  6450--6459 (2018)

\bibitem{tulyakov2018mocogan}
Tulyakov, S., Liu, M.Y., Yang, X., Kautz, J.: Mocogan: Decomposing motion and
  content for video generation. In: Proceedings of the IEEE conference on
  computer vision and pattern recognition. pp. 1526--1535 (2018)

\bibitem{vlontzos2021estimating}
Vlontzos, A., Kainz, B., Gilligan-Lee, C.M.: Estimating the probabilities of
  causation via deep monotonic twin networks. arXiv:2109.01904  (2021)

\bibitem{wang2020imaginator}
Wang, Y., Bilinski, P., Bremond, F., Dantcheva, A.: Imaginator: Conditional
  spatio-temporal gan for video generation. In: Proceedings of the IEEE/CVF
  Winter Conference on Applications of Computer Vision. pp. 1160--1169 (2020)

\bibitem{yoon2018ganite}
Yoon, J., Jordon, J., Van Der~Schaar, M.: Ganite: Estimation of individualized
  treatment effects using generative adversarial nets. In: International
  Conference on Learning Representations (2018)

\end{thebibliography}

\end{document}